\documentclass[10pt,twocolumn,letterpaper]{article}

\usepackage{iccv}
\usepackage{times}
\usepackage{epsfig}
\usepackage{graphicx}
\usepackage{amsmath}
\usepackage{amssymb}
\usepackage{comment}
\usepackage[subrefformat=parens]{subcaption}

\usepackage{enumitem}

\usepackage{url}
\usepackage{hyperref}
\usepackage[hyphenbreaks]{breakurl}

\iccvfinalcopy 

\def\iccvPaperID{****} 

\ificcvfinal\fi

\begin{document}

\title{Constructing Image--Text Pair Dataset from Books}

\author{\parbox[t]{.99\textwidth}{\centering
Yamato Okamoto
}\\
\parbox[t]{.27\textwidth}{\centering
NAVER Cloud Corp.\\
Gyeonggi-do, South Korea}
\parbox[t]{.33\textwidth}{\centering
WORKS MOBILE JAPAN Corp.\\
Tokyo, Japan}\\
\and
Haruto Toyonaga\\
Doshisha University\\
Kyoto, Japan\\
\and
Yoshihisa Ijiri\\
LINE Corporation.\\
Tokyo, Japan\\
\and
Hirokatsu Kataoka\\
LINE Corporation.\\
Tokyo, Japan\\
}

\maketitle
\ificcvfinal\thispagestyle{empty}\fi

\begin{abstract}
Digital archiving is becoming widespread owing to its effectiveness in protecting valuable books and providing knowledge to many people electronically. 
In this paper, we propose a novel approach to leverage digital archives for machine learning. 
If we can fully utilize such digitized data, machine learning has the potential to uncover unknown insights and ultimately acquire knowledge autonomously, just like humans read books.
As a first step, we design a dataset construction pipeline comprising an optical character reader (OCR), an object detector, and a layout analyzer for the autonomous extraction of image--text pairs. 
In our experiments, we apply our pipeline on old photo books to construct an image--text pair dataset, showing its effectiveness in image--text retrieval and insight extraction.
\end{abstract}

\vspace{-4mm}
\section{Introduction}
Paper media are one of the greatest inventions in human history and provide clues to ancient times. Meanwhile, by its very nature, paper can be lost or damaged over time. Therefore, digital archiving is progressing worldwide to preserve the wisdom of humans recorded in books. Such data are a record of the historical, cultural, and customary activities of the time, and it is essential to analyze and exploit digitized data as the next step of archiving.

\begin{figure}[t]
\centering
\includegraphics[width=0.45\textwidth]{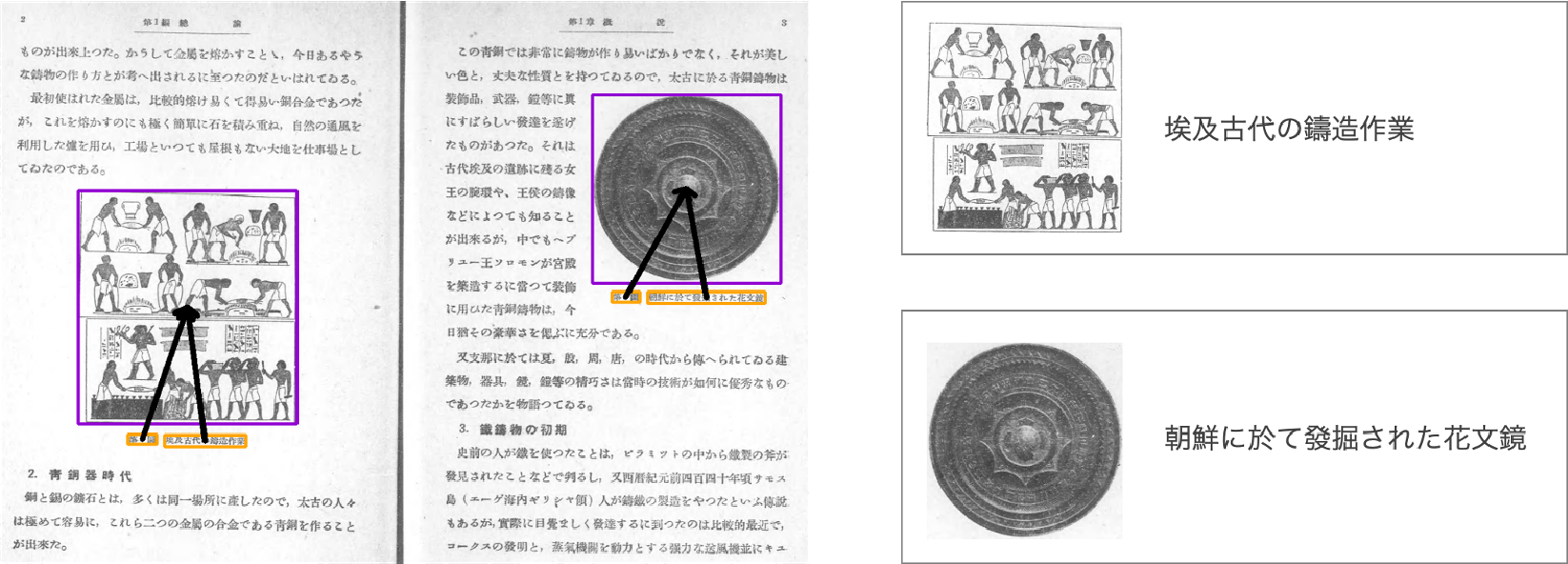}
\caption{Book images can be considered multimodal data, enabling the acquisition of image-text pairs.}
\label{image-text-pair-sample}
\end{figure}

Scanned images of books (subsequently referred to as book images) contain multimodal data, such as sentences, captions, and illustrations. By systematically linking them, we can build image-text pair datasets, allowing machine learning models to learn multimodal concepts (Fig.~\ref{image-text-pair-sample}). 
This process provides a new way to exploit digitally archived data for machine learning to acquire human-comparable knowledge autonomously, just like humans read books to learn various concepts.

Compared to sourcing data from the Internet, building datasets from books offers advantages in accuracy and may reduce the risk of the trained model outputting incorrect information. This is because published books are often meticulously fact-checked and tend to be more accurate.
Additionally, books generally include metadata, such as categories, publication year, region, and author, which can be used as labels for model training and dataset analysis.

In this paper, we propose a novel approach to leverage digital archives for machine learning. 
First, we develop a pipeline for building an image-text pair dataset from book images. This process involves extracting illustrations and their corresponding captions from book images using an optical character reader (OCR), an object detector, a layout analyzer, and a matching algorithm. It adds the metadata attached to books as supervised labels to each image and text. After developing the pipeline, we apply it to archived digital images to construct a dataset.

Next, we validate the effectiveness of training machine learning models on the dataset. In the image--text retrieval experiment, we build a retrieval system using CLIP\cite{CLIP}), and demonstrate that training with the dataset improves retrieval performance and provides new domain knowledge. 
Furthermore, in the insight extraction experiment, we build classification models using images and metadata from books and obtain unique knowledge by analyzing the model's behavior.
The contributions of this paper are summarized as follows.
\begin{itemize}[itemsep=-2pt]
    \item We develop a unique pipeline for the construction of a labeled image--text pair dataset from book images.
    \item Using the proposed pipeline, we successfully construct a practical labeled image--text pair dataset.
    \item We demonstrate a novel way of leveraging digital archives to improve image retrieval performance.   
    \item Additionally, we demonstrate the potential of digital archives in insight extraction tasks. 
\end{itemize}


\section{Related Works}
\label{Related-Work}
\paragraph{Image--Text Pair Datasets.}
A variety of downstream tasks in the vision-and-language field require image--text pair datasets, such as image--text retrieval\cite{Karpathy, Oscar}, visual question answering\cite{VQA1, VQA2, VQA3}, text-to-image generation\cite{StableDiffusion}, image captioning\cite{caption1, caption2}, and text generation from a video\cite{HashimotoAtsushi}.
As examples of large-scale image--text pair datasets, COCO Captions\cite{COCOcaption} includes over 1.5 million captions for more than 330,000 images. FashionIQ\cite{FashionIQ} includes relative text for more than 70,000 images. Flickr30k\cite{Flickr30k} contains 31,000 images with five referential sentences for each. 
The concern here is that if human annotators must assign text to images, this could become challenging and costly for larger future datasets.
\\[-6mm]

\paragraph{Dataset Construction.}
Various methods have been proposed for efficiently constructing image--text pair datasets.
Conceptual captions\cite{ConceptualCaptions} and LAION--5B\cite{LAION5b} contain 3.3 million and 5 billion image--text pairs, respectively. These pairs were obtained by crawling the Internet and matching images and their corresponding alt-texts.
FooDI--ML\cite{FooDI}, comprising more than 1.5 million images with more than 9.5 million texts related to food, was built by matching visual and text concepts via Glovo, a food delivery app. 
In WIT\cite{wikipedia}, Wikipedia was used as a data resource, and image--text pairs were extracted by using the structure of the website as clues.
Our work is in the same family as theirs but is unique in utilizing digitally archived book images as information resources.
\\[-6mm]

\paragraph{Insight Extraction.}
There are some works adopting machine learning to extract new insights from datasets. Doersch\cite{Paris} used large-scale street images provided by Google Street View. Through analysis of the model's behavior during inference cities, they discovered factors that shape Paris' identity. 
Zhou\cite{Geo} analyzed a large number of regionally labeled images and found the similarity of cities, and tried to utilize these analyses in urban planning.
Like these works, we extract new knowledge from book images as described in Section~\ref{experiments}.

\section{Dataset Construction from Book Images}

\paragraph{(i) Creation of Book Images by Digital Archives.}
\label{dataset-difficulty}
Generally, the digital archiving procedure includes scanning and calibration. It should be noted that these images often have low resolution, can sometimes be in grayscale, and may exhibit mixed text directions, including left-to-right, right-to-left, and top-to-bottom. Therefore, directly applying publicly available trained machine learning models to these images without fine-tuning can present significant challenges.
\\[-6mm]

\begin{figure}[t]
\begin{center}
\includegraphics[width=0.40\textwidth]{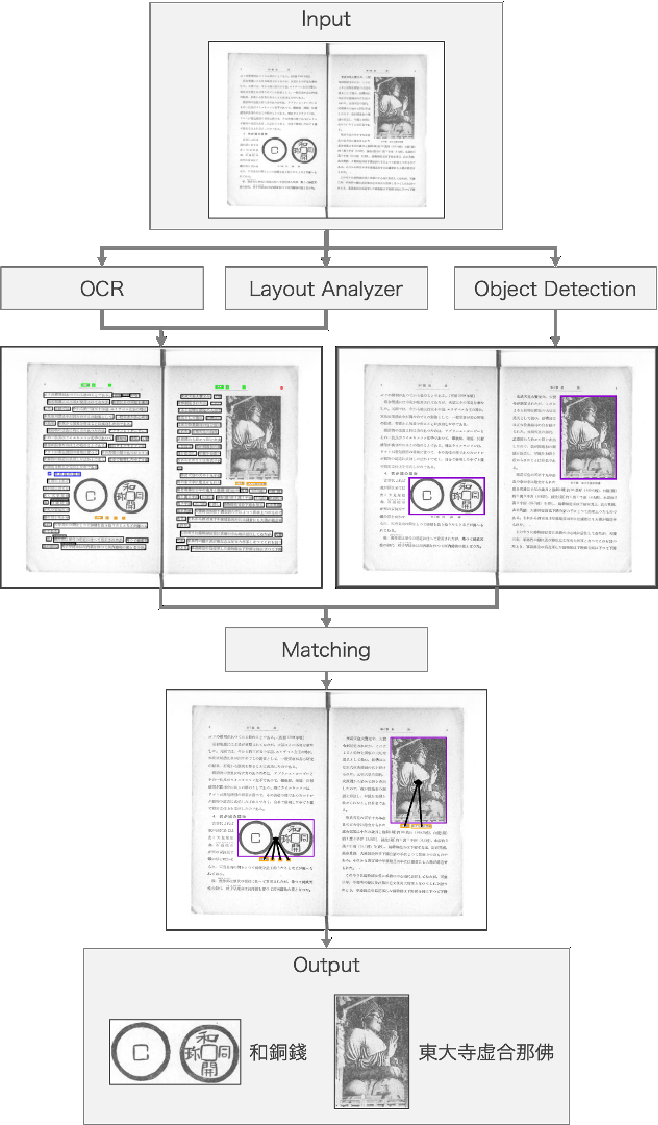}
\caption{The overview of the proposed image--text pair dataset construction pipeline.}
\vspace{-4mm}
\label{overview}
\end{center}
\end{figure}

\paragraph{(ii) Extraction of Image--Text Pairs from Book Images.}
Figure~\ref{overview} shows the overview of the proposed pipeline. 
We use an object detection model and OCR\cite{CRAFT, Recognizer} to detect illustration areas and caption areas within book images, respectively. In extracting image--text pairs from book images, illustrations should be paired with text that directly describes them (i.e., the caption). But OCR detects all text in book images, including non-caption elements. So, we use a semantic segmentation model as a layout analyzer to filter and keep only caption elements from the OCR results. 

\label{dataset-generation}

To adapt these models to the domain of digital archive images, we prepare a dataset and fine-tune each model.
The dataset is among those that can be provided by the National Diet Library (NDL)\cite{NDLｌab}, including over 10,000 digitally archived images from various genres such as literature, science, photo book, and magazines. The dataset also has labels annotated manually, indicating the regions of illustrations, bounding boxes and string labels for text regions, as well as labels for eight layout classes of book elements\footnote{Namely, captions, headlines, main text, page numbers, text within figures, notes, headers, and "don't care" background areas.}.
In practice, we compared several popular methods in semantic segmentation\cite{DeepLab, Swin, SegFormer} and object detection\cite{Cascade, MaskRCNN, DETR}, and chose SegFormer\cite{SegFormer} and Mask-RCNN\cite{MaskRCNN} which showed slightly better performance.

We obtain image--text pairs by matching captions and illustration areas. Note that a single book image may contain multiple illustrations and captions. We, therefore, link each caption to its nearest-neighbor illustration area.
\\[-6mm]

\begin{figure}[tbp]
    \centering
    \begin{tabular}{cc}
        \includegraphics[width=0.20\textwidth]{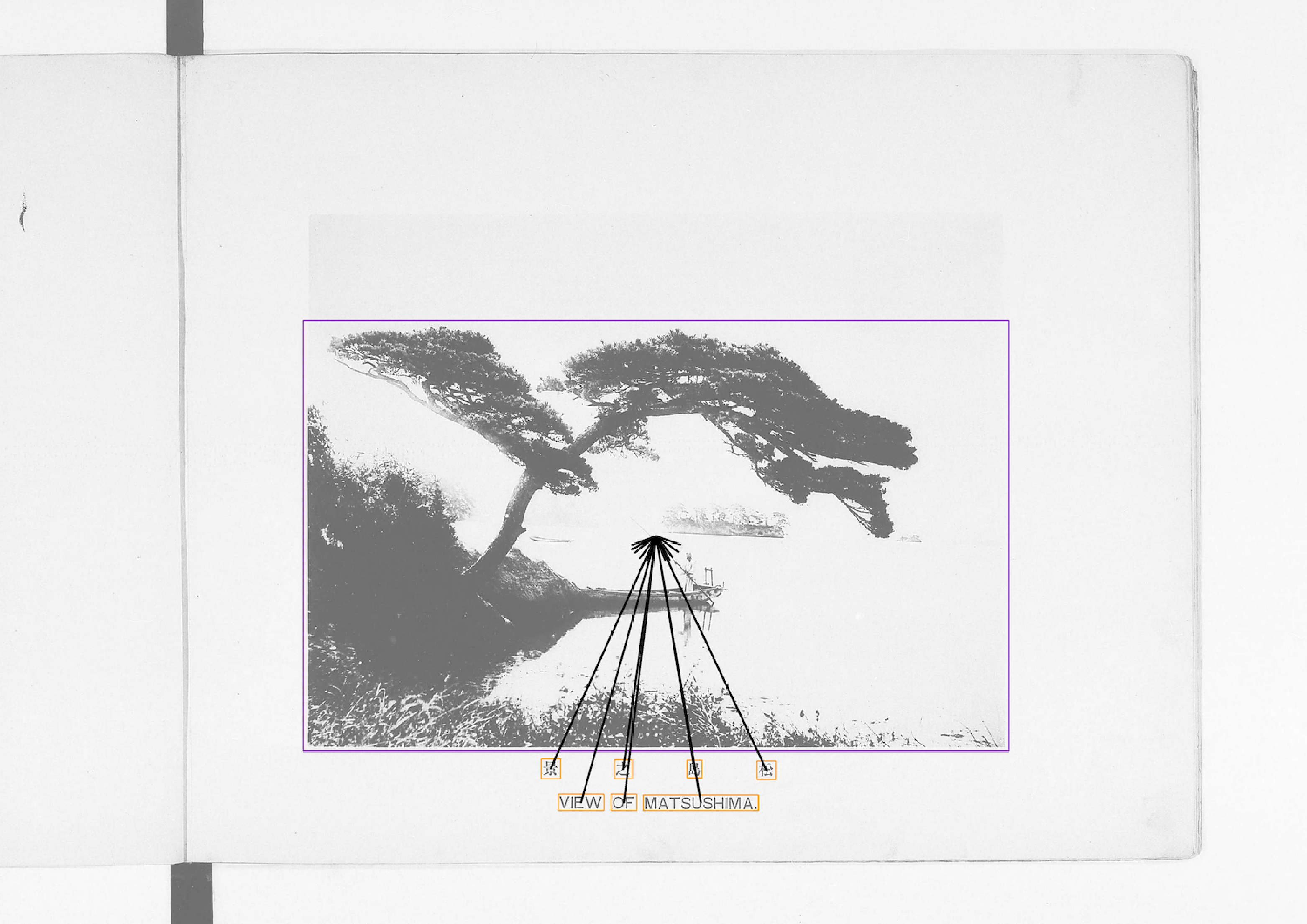} &
        \includegraphics[width=0.18\textwidth]{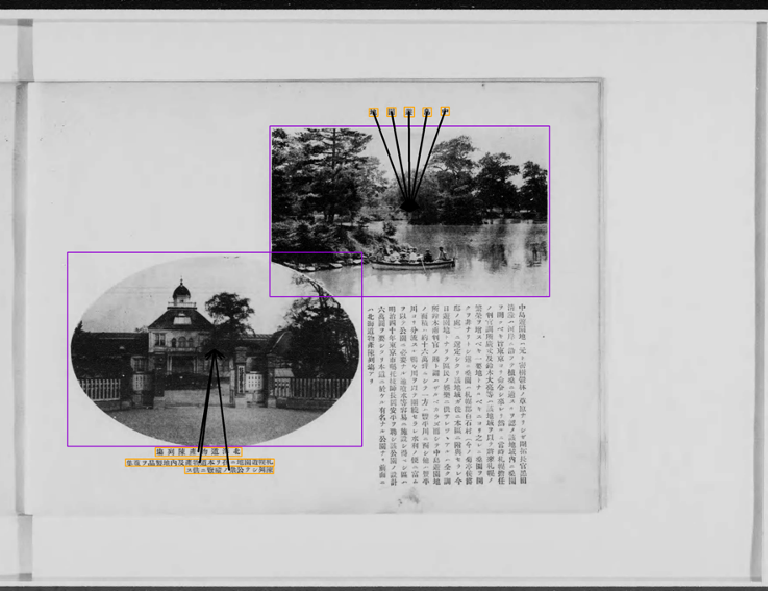}
    \end{tabular}
    \caption{Examples of book images that we used and the results of matching illustration areas and captions.}
    \label{fig:dataset-sample}
\end{figure}

\paragraph{(iii) Post Processing.}
The object detector sometimes mistakenly detects graphs and tables as illustration areas due to their decorative elements. To rectify this, we exclude graphs and tables during post-processing. If the OCR detects an amount of text surpassing a predefined threshold within a detected illustration area, assuming it is likely to be graphs or tables, that area is excluded from further processes.

linking captions to the nearest-neighbor illustration might lead to several captions being associated with a single illustration. This can occur when the caption string includes line breaks or spaces, causing the OCR to split it into multiple caption areas and output multiple detection results as shown in Figure~\ref{fig:dataset-sample}.
In this case, we combine the caption strings that are linked to the same illustration area in the order from top-left to bottom-right.

In some instances, there may be illustration areas without accompanying captions or captions that merely provide an illustration number rather than describing the content of the illustration. To prevent the creation of incomplete image--text pairs, if a caption is not associated with an illustration area or if the character count of the associated caption is below a certain threshold, it is discarded.
\\[-6mm]

\paragraph{(iv) Evaluation of The Dataset Construction Pipeline.}
The OCR and the illustration area detection showed good performance as detailed in Table~\ref{pipeline-eval}. The layout analyzer also showed good mean accuracy, however, the accuracy for the caption class alone was relatively low. Notably, there were differences in accuracy depending on the book genres. 
It performed accurately for literature, science, and photo book. Especially for photo book where captions were usually located near illustrations like Fig~\ref{fig:dataset-sample}, there were almost no errors.
Conversely, for images with complex layouts, like magazines where captions and illustrations either overlap or are separated by a distance, there were more errors.
Applying our pipelines to such complex layouts is one of the future challenges. Nevertheless, this limitation does not completely negate the utility of our pipeline.

\begin{table}[hbtp]
  \caption{Evaluation results for each pipeline component.}
  \label{pipeline-eval}
  \centering
  \small
  \begin{tabular}{|l|c|c|}
    \hline
    \multicolumn{1}{|c|}{Task} & Metric & Score \\    
    \hline
    OCR  & F-score & 0.96 \\
    \hline
    Illustration Area Detection & mAP & 0.85 \\
    \hline
    Layout Analysis (mean of eight classes) & mACC & 0.81 \\
    \hline    
    Layout Analysis (for the caption class) & ACC & 0.57 \\
    \hline
  \end{tabular}
\end{table}
\vspace{-4mm}

\paragraph{(v) Automatic Assignment of Supervised Labels.}
In most cases, digitally archived books have some kind of metadata. In our pipeline, metadata is automatically assigned as supervised labels to each extracted image and text.

\section{Experiment}
\label{experiments}

\subsection{Dataset}
We experimented with digital archives published by the National Diet Library. We used the photo books\cite{NDLphoto} published around 1868--1945 containing a large number of photographs and descriptions of landscape and historical buildings of almost all regions (called prefectures) in Japan. 

Photo books can be considered suitable for validating the proposed approach to leverage digital archiving for machine learning, as our dataset construction pipeline demonstrated particularly accurate behavior for photo books. 
The details of these digital archives are described in \cite{NDLArchiveMethod}. We used 175 photo books containing 12,640 book images and obtained 9516 image--text pairs. Each photo book had prefecture metadata, and we assigned prefecture labels to each pair.

\subsection{Experiment 1: Image--Text Retrieval}
\label{experiments2}
With the autonomously built image--text pair dataset, we conducted image--text retrieval experiments to show the effectiveness of training the dataset. We first constructed a cross-modal retrieval system using CLIP. This system extracts features from a text query and acquire images with relevant features from the dataset. 
We then split the image--text pair dataset into 8016 training data and 1500 testing data and trained the CLIP with the training data for 20 epochs, using the ViT--B/32 publicly available backbone for initialization and the default values for all hyperparameters. 
We evaluated the CLIP performance 10 times and calculated the average recall, using 1,000 image--text pairs randomly selected from the test data.

\begin{figure}[htb]
    \centering
    \begin{tabular}{|c|c|}
      \hline
      \raisebox{5mm}{ViT--B/32} 
      & \includegraphics[keepaspectratio, scale=0.245]{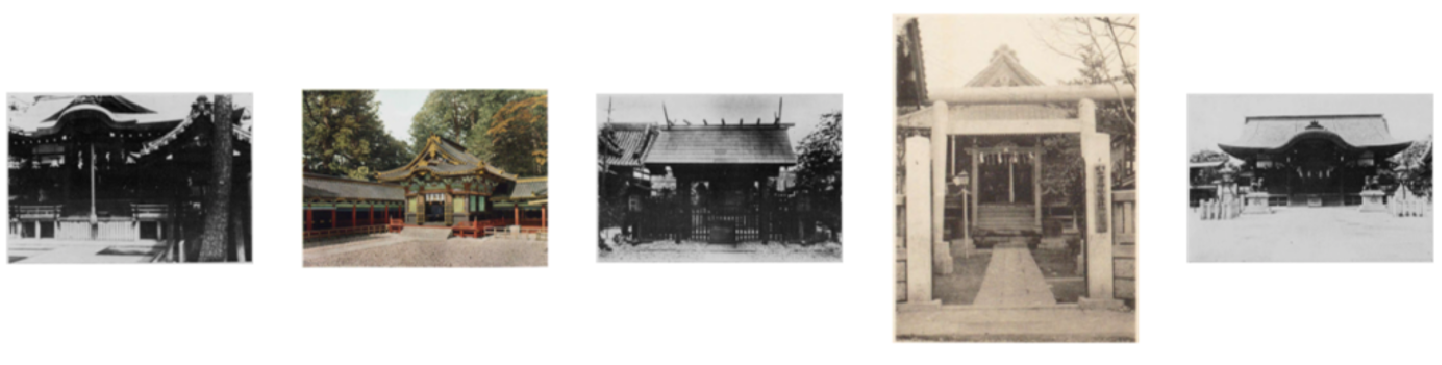} \\
      \hline
      \raisebox{5mm}{Ours}  
      & \includegraphics[keepaspectratio, scale=0.245]{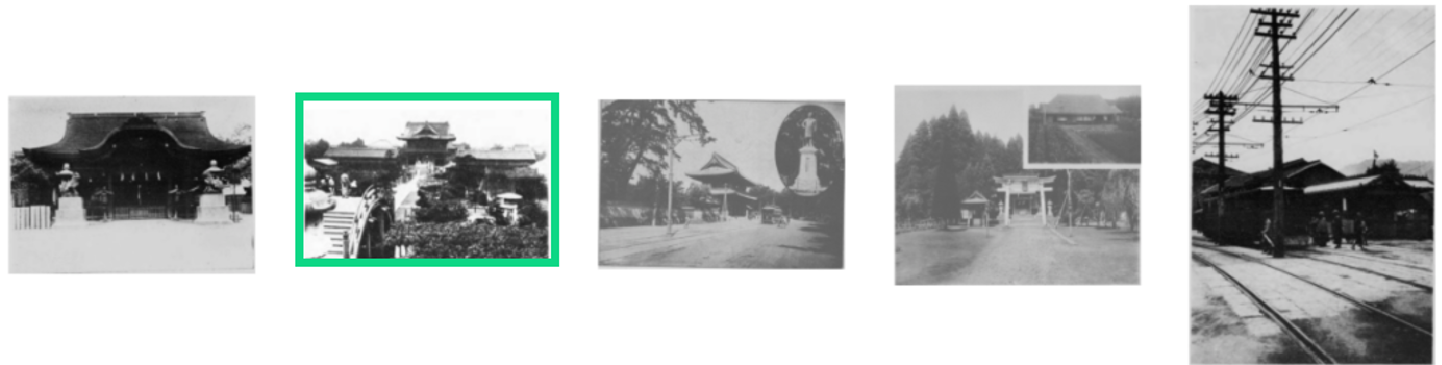} \\
      \hline 
    \end{tabular}
  \caption{
  The top five images retrieved with text containing "KAMEIDO TENMANGU" which is the name of the shrine in Japan. In this way, the model trained with Japanese photo books could retrieve the images of a specific Japanese location or build through query text of their name (green bounding boxes indicate correct images).
  }
  \label{retrieval}
\end{figure}

Table~\ref{table:retrieval} presents the evaluation results. Training the model on the dataset improved the retrieval performance. Moreover, there were instances where the model retrieved images of specific Japanese locations or buildings based on query texts with their names, as shown in Figure~\ref{retrieval}. These results imply that the dataset built from digital archives provides new domain knowledge to CLIP.
%
\begin{table}[htb]
  \caption{Evaluation of retrieval performance with Recall@K (1k images and 1k texts).}
  \label{table:retrieval}
  \centering
  \small
  \begin{tabular}{|c|c|c|c|c|}
    \hline
       & \multicolumn{2}{c|}{Image2Text Retrieval} & \multicolumn{2}{c|}{Text2Image Retrieval} \\
    \hline
       & ViT--B/32 & Ours & ViT--B/32 & Ours \\
    \hline
    Recall@1   & 0.01 & 0.04 & 0.01 & 0.05 \\
    Recall@10  & 0.04 & 0.16 & 0.03 & 0.16 \\
    Recall@50  & 0.12 & 0.38 & 0.12 & 0.39 \\
    Recall@100 & 0.20 & 0.51 & 0.20 & 0.53 \\
    \hline
  \end{tabular}
\end{table}
\vspace{-3mm}

\subsection{Experiment 2: Insight Extraction}
\label{experiments1}
In this experiment, our objective was to extract insights from digital archives.
First, we a built Japanese prefecture classification model by training images and prefecture labels. the model took images as input and simply consisted of convolution layers and a fully connected layer. We split the dataset at a ratio of 9:1 into training and testing data. After training for 20 epochs, the model's accuracy reached 0.76.
By analyzing the behavior of the models, we extracted unique insights related to Japanese prefectures.
\\[-6mm]

\begin{figure}[htb]
\centering
\includegraphics[width=0.38\textwidth]{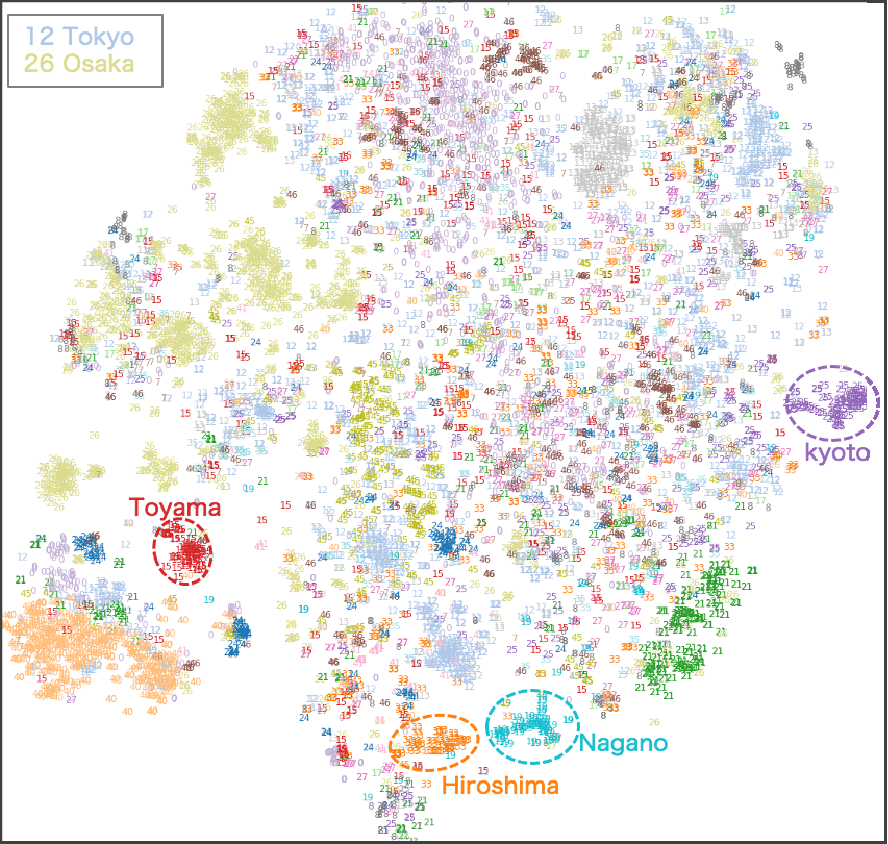}
\caption{Visualization of the feature distribution by t-SNE. Only some classes are shown for better visualization.} 
\label{fig:TSNE}
\end{figure}

\paragraph{(i) Analysis Based on the Distribution of Features.}
Figure~\ref{fig:TSNE} shows the distribution of features, visualized by t-SNE\cite{tSNE}.
The distribution revealed unexpected similarities and differences among prefectures, unrelated to geographical distance. For instance, geographically distant Nagano (classID=19) and Hiroshima (classID=33) clustered closely, while neighboring Toyama (ClassID=15) and Nagano clustered distantly.
The distribution also showed reasonable tendencies. Kyoto (classID=25), a unique city with many historical buildings, formed one cluster without mixing with other prefectures.

\vspace{-4mm}

\begin{figure}[htb]
    \centering
    \begin{tabular}{cc}
        \includegraphics[keepaspectratio, scale=0.155]{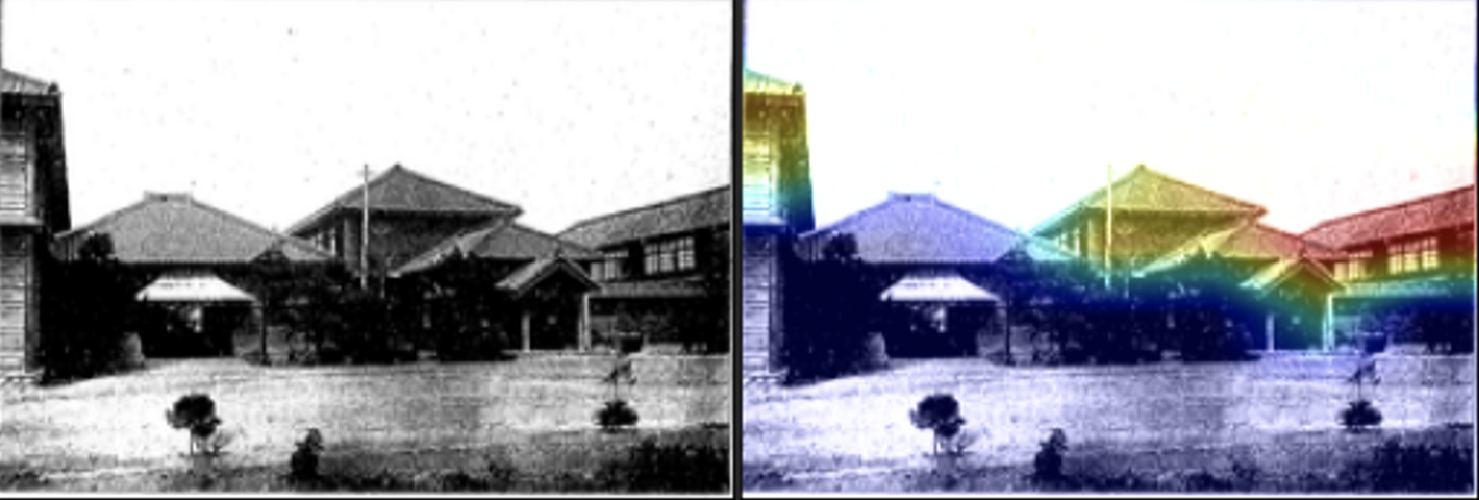} &
        \includegraphics[keepaspectratio, scale=0.155]{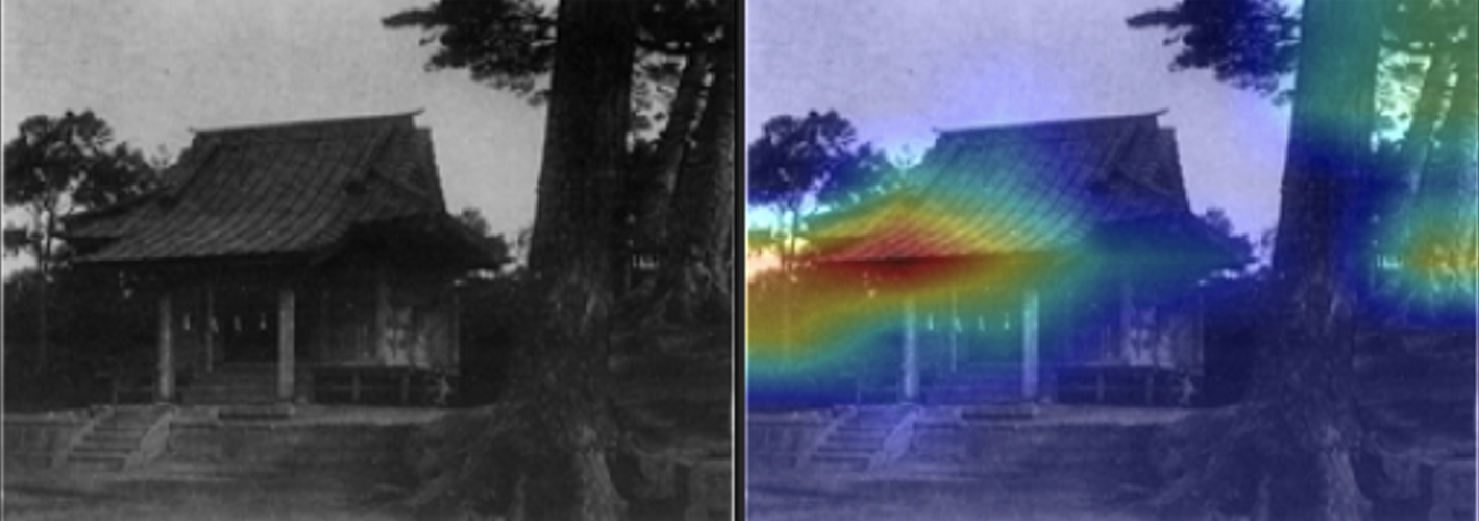}
    \end{tabular}
    \vspace{-4mm}
    \caption{Grad-CAM visualization.}
    \label{fig:GradCAM}
\end{figure}

\paragraph{(ii) Analysis of Activation.}
We used Grad-CAM\cite{GradCAM} and visualized activated areas in images, likely reflecting prefecture identities. The analysis highlighted the building roof structures (Fig.~\ref{fig:GradCAM}). This is presumably because roof structures vary to accommodate regional climates.

\vspace{-4mm}

\section{Conclusion}
\label{conclusion}
In this paper, we proposed a new approach to utilize digital archives by creating an image-text pair dataset.
We demonstrated the effectiveness of training on the dataset in terms of improving the retrieval performance in a specific domain and obtaining insight such as the similarities and uniqueness of the learned concept.
This is the first step to realizing machine learning to acquire knowledge autonomously, just like humans read books.

\section{Acknowledgments}
\vspace{-1mm}
We used a publicly-available digital archive and NDL-DocL Dataset 1.0\cite{NDLDocL} from the National Diet Library. We express our deepest gratitude for the use of the data.

{\small
\bibliographystyle{ieee_fullname}
\bibliography{egbib}
}

\end{document}